\documentclass{article}
\usepackage{ijcai19}

\usepackage{times}
\usepackage{soul}
\usepackage{url}
\usepackage[hidelinks]{hyperref}
\usepackage[utf8]{inputenc}
\usepackage[small]{caption}
\usepackage{graphicx}
\usepackage{amsmath}
\usepackage{stmaryrd}

\usepackage{booktabs}
\usepackage{algorithm}
\usepackage{algorithmic}

\usepackage{makecell}
\usepackage{graphicx}
\usepackage{amsmath}
\usepackage{multirow}
\usepackage{amsfonts}
\usepackage{url}
\usepackage{xcolor}
\usepackage{subcaption}
\usepackage[justification=centering]{caption}
\usepackage{float}
\title{GPS: A Policy-driven Sampling Approach for Graph Representation Learning}

\author{
Tiehua Zhang$^1$\and
Yuze Liu$^1$\and
Xin Chen$^{1}$\and
Xiaowei Huang$^{1}$\and
Feng Zhu$^{1}$\and
Xi Zheng$^2$\footnote{Corresponding Author}\\
\affiliations
$^1$Ant Group\\
$^2$Macquarie University\\
\emails
\{zhangtiehua.zth, liuyuze.liuyuze, jinming.cx, wei.huangxw, zhufeng.zhu\}@antgroup.com\\
james.zheng@mq.edu.au
}

\begin{document}

\maketitle

\begin{abstract}
Graph representation learning has drawn increasing attention in recent years, especially for learning the low dimensional embedding at both node and graph level for classification and recommendation tasks. To enable learning the representation on the large-scale graph data in the real world, numerous research has focused on developing different sampling strategies to facilitate the training process. Herein, we propose an adaptive \textbf{G}raph \textbf{P}olicy-driven \textbf{S}ampling model (GPS), where the influence of each node in the local neighborhood is realized through the adaptive correlation calculation. Specifically, the selections of the neighbors are guided by an adaptive policy algorithm, contributing directly to the message aggregation, node embedding updating, and graph level readout steps. We then conduct comprehensive experiments against baseline methods on graph classification tasks from various perspectives. Our proposed model outperforms the existing ones by 3\%-8\% on several vital benchmarks, achieving state-of-the-art performance in real-world datasets.
\end{abstract}

\section{Introduction}
Graph neural networks (GNN) are now the \textit{de facto} standard when it comes to learning the low-dimension embedding on the graph-structured data. It is shown effective in prior works to use the low-dimensional vector embeddings as the input node/graph features for a variety of prediction, recommendation, or any other graph-related analysis~\cite{wu2020comprehensive}. The idea behind it is through aggregating and distilling the high-dimensional information of a node as well as its graph neighborhood into a dense vector form, the principle of which is derived from the natural languages processing (NLP) field~\cite{zhou2020graph}. 

The rise of GNNs comes from the fact that most dominant neural network architectures in the deep learning field focus on dealing with tasks involving grid-like data structures (e.g., images, text corpus). However, it is unpractical not to consider the sheer amount of non-euclidean space data in the case of social networks, molecular protein networks, or financial risk networks~\cite{zhou2020graph}. These types of data can be formed into the graph structure and convey a better interpretation from a human's perspective.

Much interesting research has thus emerged over the last several years to embed the graph-structured data using GNNs, and these GNNs can be generalized in two forms, namely conventional and sampling-based ones~\cite{wu2020comprehensive}. Specifically, a conventional GNN refers to optimizing node's embedding using matrix factorization-based objective in a single, fixed graph, while aggregating knowledge from all connected node pairs. One widely used work is graph convolutional networks (GCN), which could easily lead to the ``neighbor explosion'' problem in the large-scale and dense graph. The recursive expansion across layers in a top-down way also requires expensive convolution computations for all nodes at each time step. One flagship sampling-based model is GraphSAGE, which randomly samples a fixed size of neighbors for each node instead of taking in the whole neighborhood to avoid ``neighbor explosion'' problem. Similarly, other sampling works either resort to the layer-wise sampling~\cite{huang2018adaptive} or graph-level sampling~\cite{graphsaint}. 



It is considered challenging for representation learning due to the need of generalizing the ones on which the algorithm has already been optimized to the unseen nodes/graphs. To this end, we propose an adaptive graph policy-driven sampling model, namely GPS, to generate both node and graph level embedding inductively. The policy-driven sampling is realized through several key steps. It first quantifies the importance of all nodes at each layer recursively, from which a better sense of locality of a given node can be comprehended while sampling the nodes based on the rank of importance. It then integrates the sampling loss into the final graph level classification task, and the importance of the nodes is thus adjusted adaptively along with the change of gradients towards the minimum of loss. The contributions of the paper can be summarized as the following:
\begin{itemize}
\item We propose an innovative policy-driven sampling approach to enhance the generation of the embedding on the graph-structured data, in which the node embedding is updated with the guidance of the policy function, and the graph embedding derives from a shallow multi-layer perceptions network (MLP).
\item We realize the graph representation learning in an end-to-end fashion, including the initial graph construction/vectorization, the adaptive policy-driven sampling/embedding update and graph embedding generation. We quantify the correlations of nodes in the local neighborhood and only sample ones deemed important iteratively. In addition, the sampling process is considered an integral part of the training process and thus adapted dynamically.
\item We conduct extensive experiments on five molecular graph datasets for graph level classification tasks. Compared with the other four state-of-the-art methods, the proposed GPS achieves superior performance among all. We also demonstrate the influence of sampling parameter $K$ on the overall performance of the GPS and implement variants of GPS for ablation study.
\end{itemize}
\begin{figure*}[h!]
    \vspace{-6mm}
    \centering
    \includegraphics[width =\textwidth]{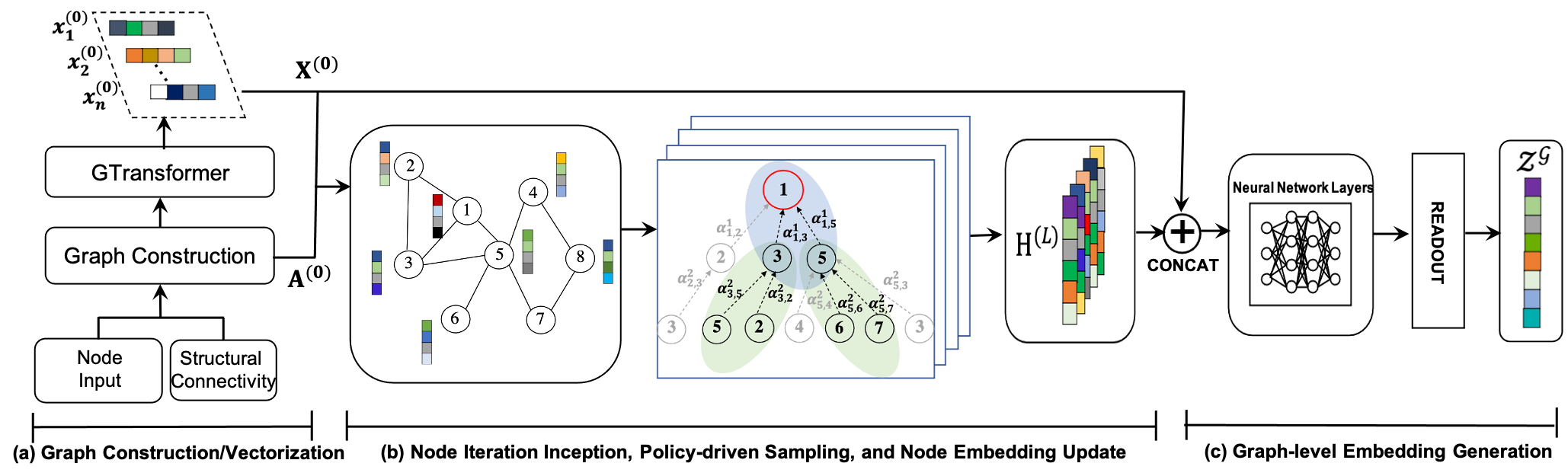}
    \vspace{-6mm}
    \caption{The overview of node feature vectorization, and embedding update via GPS, and graph-level embedding generation (from left to right).}
    \label{fig:demo}
    \vspace{-6mm}
\end{figure*}
\section{Related Work}
While the sheer number of graph-structured data have been found particularly useful in many learning tasks in recent years (e.g., financial risks analysis for costumers social connection graphs), designing effective and efficient graph neural networks has become a popular topic in the graph representation learning field. 

\cite{bruna2013spectral} first introduces the convolution operation into the graph domain and propose both spatial construction and spectral construction during the convolution process. To speed up the graph convolution computation, both \cite{gcn} and \cite{defferrard2016convolutional} design localized filters based on Chebyshev expansion and verify their approach on relatively small datasets, where the training is able to happen in full batch. Apart from that, \cite{duvenaud2015convolutional} adopts the convolution by using the spatial connections on the graph directly, where a wight matrix is learned for each node degree. Additionally, \cite{cao2019multi} develops the channel-wise convolution to learn a series of graph channels at each
layer, which is able to shrink the graph hierarchically to encode the pooled structures. These GCN-based models have achieved state-of-the-art performance at that time and outperformed the traditional node embedding approaches such as node2vec~\cite{grover2016node2vec} and DeepWalk~\cite{perozzi2014deepwalk}.

However, despite the success of GCN-based models, one major limitation is that these models are all based on knowledge of the entire graph Laplacian matrix amid training, making it unpractical to large-scale graph data with millions of nodes and edges. Following that, various sampling approaches are proposed to avoid operating on the entire Laplacian matrix~\cite{graphsage,ying2018graph,huang2018adaptive,graphsaint}. Specifically, \cite{graphsage} first introduces the uniform layer-wise sampling algorithm, which applies a fixed size sampling window to each layer from inside out iteratively. \cite{ying2018graph} takes advantage of random walk and considers sampling from different perspective. It considers nodes with highest visited count as important nodes and constructs the neighborhood based on this. Both~\cite{huang2018adaptive} and~\cite{graphsaint} focus on the variance reduction-based sampling. While the former proposes a condition-based sampling where the lower layer is sampled conditionally on the top one, the latter samples at the graph level, suggesting to construct different subgraphs as the minibatches while decoupling the sampling from the forward and backpropogation processes.

Another line of work aims to improve the model capacity. For instance, \cite{gat} first integrates the attention mechanism into the graph representation learning. They introduce an attention-based architecture and the use self-attention to assign edge weights to the neighborhood. Following that, \cite{zhang2018gaan} adopts gated attention network to further enhance the control over attention heads. While our proposed method shares the same concept of quantifying the correlations to the neighbor nodes, it differentiates in coefficients calculations, the adaptive policy-driven sampling for neighbor nodes, and the forward message passing process. 

\section{The Proposed GPS Model}
The target of graph classification task is to map a structural graph data into a set of labels. In this work, we consider a full undirected graph
$\mathcal{G} = \left(\mathcal{V}, \mathcal{E}\right)$, where 
$\mathcal{V} = \left\{\emph{v}_{1},...,\emph{v}_N\right\}$ is the set of nodes, and $\mathcal{E} = \left\{\left(\emph{v}_{i},\emph{v}_{j}\right)|\emph{v}_{i},\emph{v}_{j}\in\mathcal{V}\right\}$ denotes the set of edges in the graph. We use an adjacency matrix $\emph{A}\in\left\{0,1\right\}^{N\times N}$ to encompass the edge sets with all connectivity information in which $\emph{A}_{i,j} = 1$ if $\left(\emph{v}_{i},\emph{v}_{j}\right)\in\mathcal{E}$ and $\emph{A}_{i,j} = 0$ otherwise. Since GNN takes both adjacency matrix and vectorized node feature as input to learn the node embedding, we herein have initial node features denoted by $\emph{X}^{0} \in \mathcal{R}^{N\times d}$ with $\emph{x}_{i} \in X^0$ corresponding to the $d$-dimensional feature vector for node $\emph{v}_i$. Simply, the neighbor nodes of $\emph{v}_{i}$ are represented as $\mathcal{N}\left(\emph{v}_{i}\right)$. For the graph classification task, we use $\left\{\mathcal{G}_{1},...,\mathcal{G}_{n}\right\}$ as the whole graph datasets and $\left\{\emph{y}_{1},...,\emph{y}_{n}\right\}$ as corresponding labels, before splitting into train/val/test set.
\paragraph{Solution to Initial Node Embedding}GNN requires both vectorized feature input and adjacency matrix~\cite{yuan2020explainability,rong2019dropedge}, yet original node features in the real-world structural graphs are not vectorized in the first place (e.g., citation network, molecular graph, and social network)~\cite{lou2021stfl}. To this end, we demonstrate the process of turning raw node feature input to initial node embeddings in Figure~\ref{fig:demo}(a), including the steps of graph construction as well as vectorization. At the graph construction step, the graph is assembled with both raw node feature (e.g., paper/author name in citation graph, or carbon atom name in molecular graph) and structural connectivity (i.e., a list of pair-wise connections using node ids), transforming the graph into sequential representation known as SMILES format~\cite{weininger1989smiles}. It is not uncommon to use SMILES format to encode the contextual information of the graph on which the NLP-based models can be easily applied to generate the node embeddings~\cite{wang2019smiles,liu2019n}. To better encode the structural information of the graphs that other NLP-based models fail to do, we use a pre-trained GNN Transformer called GTransformer to leverage both the node features and structural information in SMILES sequence when it comes to generating the initial feature embeddings, which is then used as the input to GPS. Briefly, the key component of the GTransformer is the graph multi-head attention component, which is the attention blocks tailored to latent structural data and reports the state-of-the-art initial feature embedding results. The detailed model architecture can be found in~\cite{rong2020self}.

\paragraph{Embedding Update and Learning}Instead of drawing the full expansion of the graph like GCN~\cite{gcn} in the feedforward computation, our method updates the embedding of each node iteratively (Figure~\ref{fig:demo}(b)). Specifically, we define the sampled node set as $\emph{V}_{sa}^{l}$ ($\emph{V}_{sa}^{l} \subseteq \mathcal{V}$) whose embedding only need to be updated at aggregation layer $l$. Note $\emph{V}_{sa}^{0}$ essentially contains all nodes at the root layer, while sampled nodes at other layers are adaptively chosen through the designed policy-driven sampling process (details unveiled in eq.(4) and Algorithm 1). The unduplicated neighbor set of $\emph{V}_{sa}^{l}$  is denoted as $\emph{V}_{ne}^{l}$ = $\left\{\mathcal{N}\left(\emph{v}_{i}^{l}\right) : \forall\emph{v}_{i}^{l}\in\emph{V}_{sa}^{l} \right\}$ in which $\mathcal{N}\left(\cdot\right)$ returns the neighbor nodes of ${v}_{i}^{l}$. To show a concrete example, Figure~\ref{fig:demo}(b) demonstrates one of the sampled nodes $v_1^0$ at the root layer ($l=0$) with neighbor node set \{$v_2^1$, $v_3^1$, $v_5^1$\}. We use words aggregation layer and message passing layer interchangeably in this paper, the notation of which is $l \in \left\{1,...,L\right\}$, and $L$ indicates the largest aggregation/message passing layer.

Since the node index are in fixed order in full adjacency matrix $\emph{A}$, we can define both row filter matrix $\emph{R}^{l}_{sa} \in \mathcal{R}^{\left\|\emph{V}_{sa}^{l}\right\|\times N}$ and column filter matrix $\emph{C}^{l}_{ne} \in \mathcal{R}^{N \times \left\|\emph{V}_{ne}^{l}\right\|}$ to select rows or columns from the adjacent matrix at any message passing layer $l$. Simply put, the left multiplication with $\emph{R}^{l}_{sa}$ can be regarded as selecting sampled nodes from full adjacency $\emph{A}$, and the right multiplication with $\emph{C}^{l}_{ne}$ indicates selecting corresponding neighbor nodes. Taking the row filter matrix as an example, $\emph{R}^{l}_{sa}$ is expanded as $\left[\emph{e}_{1}^l,\emph{e}_{2}^l,...,\emph{e}_{i}^l \right]^{T}$, where $\emph{e}_{i} \in \{0, 1\}^{N \times 1}$ and $i=\left\|\emph{V}_{sa}^{l}\right\|$. Specifically, $\emph{e}_{i}^l$ serves as the position indicator and tells the $i$th sampled node position at layer $l$ in the full adjacency matrix $A$, e.g., $\emph{e}_{1}^l = \left[0,1,0,...,0\right]^T \in \mathcal{R}^{N \times 1}$ means the first sampled node is at the second row in $\emph{A}$ (node 2). Herein, the column filter matrix $\emph{C}^{l}_{ne}$ follows the same logic to filter out the unique neighbor nodes of the sampled nodes. Consequently, the layer-wise adjacency matrix $\emph{A}^{l} \in \mathcal{R}^{\left\|\emph{V}_{sa}^{l}\right\|\times \left\|\emph{V}_{ne}^{l}\right\|}$ can be formulated as:


\begin{equation}
    \emph{A}^{l} = \left(\emph{R}^{l}_{sa}\cdot\emph{A}\right)\cdot\emph{C}^{l}_{ne} 
\end{equation}


Similarly, we can retrieve the embedding representation for both sampled nodes and neighbor nodes at layer $l$. Say the hidden embedding matrix of all nodes at hop layer $l$ is represented as $\emph{H}^{l} \in \mathcal{R}^{N\times D}$, and $\emph{H}^{0} = \emph{X}^0$ simply means the node embedding is assigned with the initial node feature matrix at the root layer. 

\begin{equation}
    \hat{\emph{H}}^{l}_{sa} = \emph{R}^{l}_{sa}\cdot\emph{H}^{l}
\end{equation}

\begin{equation}
   \tilde{\emph{H}}^{l}_{ne} = \left(\emph{C}^{l}_{ne}\right)^{T}\cdot\emph{H}^{l}
\end{equation}
where $\hat{\emph{H}}^{l}_{sa}\in\mathcal{R}^{\|\emph{V}_{sa}^{l}\|\times D}$ and $\tilde{\emph{H}}^{l}_{ne}\in\mathcal{R}^{\|\emph{V}_{ne}^{l}\|\times D}$ represent sampled nodes embedding and unique neighbor nodes, respectively.

We then calculate the correlation coefficients for each connected node pair and proceed with the sampling process, which is an integral part of the policy strategy in the proposed model. The adaptive policy-driven sampling process can be written as:

\begin{equation}
    \emph{A}^{l}_{\alpha}= \Phi\left[\delta\left(\hat{\emph{H}}^{l}_{sa}\cdot\emph{W}^{l}_{sa}, \left(\tilde{\emph{H}}^{l}_{ne}\cdot\emph{W}^{l}_{ne}\right)^{T}\right)\odot\emph{A}^{l}\right]
\end{equation}

$\delta\left(\cdot\right)$ denotes the coefficient function, which is used to quantify the correlations of each sampled node with different neighbor nodes. Instead of applying coefficient function directly on sampled node embedding $\hat{\emph{H}}^{l}_{sa}$ and neighbor node embedding $\tilde{\emph{H}}^{l}_{ne}$, we utilize two learnable weight matrices $\emph{W}^{l}_{sa}$ and $\emph{W}^{l}_{ne}$ to conduct further linear transformations in which embedding space for both sampled and neighbor nodes can be approximated separately at that layer throughout the training. Towards optimizing two transformation matrices for both sampled and neighbor node embeddings, our model can adaptively uncover better correlations through designated operations (e.g., multiplicative, normalized cosine or distance based operations, etc.) in the transformed embedding space. The Hadamard product operator $\odot\left(\cdot\right)$ is employed to only retain the connected node pairs. $\Phi\left(\cdot\right)$ is the policy function that confines a pre-defined budget on the
sample size, from which the computation complexity can be bound easily. In this paper, we implement $\emph{top}\left(K,\cdot\right)$ in row-wise fashion as the policy to only selects neighbors deemed important in this case, and $K$ is the sample budget. $\emph{A}^{l}_{\alpha}$ is the calculated correlation matrix at that layer, and $\alpha_{i, j}^l \in \emph{A}^{l}_{\alpha}$ has also been visually shown in Figure~\ref{fig:demo}(b).

Following that, we have $\hat{\emph{h}}^{l}_{i}\in\hat{\emph{H}}^{l}_{sa}$ to represent embedding of each sampled node, where $i\in\{1,...,\|\emph{V}_{sa}^{l}\|\}$. The message aggregation from layer $l$ to $l+1$ can be formalized as:

\begin{equation}
    \hat{\emph{h}}_{i}^{l+1} = \sigma\left(\emph{W}^{l+1}_{agg}\cdot\left(\hat{\emph{h}}_{i}^{l}\oplus\frac{1}{\|\emph{V}_{ne}^{l}\|}\sum_{j=1,..,\|\emph{V}_{ne}^{l}\|}\alpha_{i,j}^{l}\tilde{\emph{h}}_{j}^{l}\right)\right)
\end{equation}
where $\tilde{\emph{h}}^{l}_{j}\in\tilde{\emph{H}}^{l}_{ne}$, $\oplus$ denotes the concatenation operator, and $\sigma$ is the nonlinear activation function (e.g., ReLu). Note we use $\|\emph{V}_{ne}^{l}\|$ as the regularizer after aggregating messages from neighbor nodes. $\emph{W}^{l+1}_{agg}$ is a shared weight matrix at that layer and applied on every sampled node before updating its embedding.

\paragraph{Graph-level Representation}To derive the final graph representation from the node embeddings, we design the \textit{READOUT} operation as follows:  

\begin{equation}
    \emph{z}^{\mathcal{G}} = MEAN\left(MLP\left(\emph{X}^{0}\oplus\emph{H}^{L}\right)\right)
\end{equation}

Instead of simply taking the mean of all node embeddings, we first input the embedding matrix to a shallow multi-layer perception (MLP), the design of this MLP could be found in the experimental setup in Section 4. The row-wise $MEAN$ operation is then applied on the output of the MLP, as shown in Figure~\ref{fig:demo}(c). For notational convenience, we use $\hat{\emph{h}}_{i}^{L} \in H^L$ to denote the final node representations.

In order to learn graph representation in the fully supervised setting, we apply a graph-based cross entropy loss onto the graph-level output $\emph{z}^{\mathcal{G}}$. The gradient-based adjustments for all trainable parameters mentioned above achieve two goals: 1) to select the neighbor nodes with the best interests adaptively; 2) to approach the suitable latent embedding space gradually.

\begin{equation}
    \mathcal{L}\left(\emph{z}^{\mathcal{G}},\emph{y}^{\mathcal{G}}\right) = -\sum_{i}\left[\emph{y}^{\mathcal{G}}_{i}\cdot\log\emph{z}^{\mathcal{G}}_{i} +  \left(1-\emph{y}_{i}^{\mathcal{G}}\right)\cdot\log\left(1-\emph{z}^{\mathcal{G}}_{i}\right)\right]
\end{equation}
where the loss $\mathcal{L}$ of graph output $\emph{z}^{\mathcal{G}}$ is calculated against corresponding label $\emph{y}^{\mathcal{G}}$, and $i$ means $i$th logits at that index.

\begin{algorithm}[t!]
	\caption{GPS training algorithm}
	\label{alg:algorith}
	\begin{algorithmic}[1]
		\REQUIRE Graph $\mathcal{G} = \left(\mathcal{V},\mathcal{E}\right)$; label $\emph{y}^{\mathcal{G}}$; sampling hyperparameter $K$; total message passing layers $L$
		\ENSURE Embedding representation of graph $z^\mathcal{G}$
		\STATE $\emph{V}_{sa}^{0}$ $\leftarrow$ root nodes $\emph{V}_{rt} \subseteq \mathcal{V}$
		\FOR{l = 1,..$L$}
		\STATE $\emph{V}_{sa}^{l}$ $\leftarrow$ Sampled nodes from previous layer $\emph{V}_{sa}^{l-1}$
		\STATE  $\emph{V}_{ne}^{l}$ $\leftarrow$ All neighbor nodes of $\emph{V}_{sa}^{l}$
		\STATE $\emph{A}^{l}$ $\leftarrow$ Generate layer-wise adjacency (eq.(1))
		\STATE $ \emph{A}^{l}_{\alpha}$ $\leftarrow$ Policy-driven adaptive sampling according to eq.(4) with adjustable parameter $K$
		\STATE $\hat{\emph{H}}^{l}_{sa}$ $\leftarrow$ Embedding update through eq.(5)
		\STATE $\emph{V}_{sa}^{l}$ $\leftarrow$ Update sampled nodes from $\emph{A}^{l}_{\alpha}$ for next layer
		\ENDFOR
		\STATE $\emph{z}^{\mathcal{G}}$ $\leftarrow$ Graph-level representation from eq.(6)
		\STATE Calculate loss from eq.(7), and update all trainable weights.
	\end{algorithmic}
	\vspace{-1.6mm}
\end{algorithm}

Algorithm~\ref{alg:algorith} illustrates the training process of one graph using GPS. There are a few pre-processing to nodes before the training starts, including the generation of sampled nodes and neighbor nodes (lines 1, 3 and 4, respectively). We want to point out that the initial sampled nodes at layer 0 comprise either a subset nodes or all nodes in the graph, depending on the size of the input graph. Line 5 constructs the layer-wise adjacency, and the sampled version of which is then generated at line 6. It can be interpreted that the algorithm only allows the neighbors deemed important to contribute to the update of embeddings of parental nodes (line 7), and the influence of the sampling parameter $K$ is studied in the next section. Note that the algorithm only passes the sampled nodes to the next-hop layer (line 8). Afterwards, the graph-level representation is composed at line 10, and all trainable weights are updated by minimizing the loss at line 11. The use of optimizer is shown in Table~\ref{tab:hyperpara}.

\paragraph{Discussion} The advantage of GPS is twofold. First, the sampling is guided by the designated adaptive sampling policy, enabling a more effective message propagation than GraphSAGE, whose node representations are generated through random sampling of neighborhoods of each node. Other importance-based sampling works adopt a different strategy, where inter-node probabilities need to be calculated and optimized to reduce the variance caused by sampled nodes~\cite{huang2018adaptive,graphsaint}. However, this line of work incurs a significant computation overhead in AS-GCN~\cite{huang2018adaptive}. The latest work GraphSAINT splits sampled nodes and edges from a bigger graph into subgraphs on which the GCN is applied, causing unstable performance on small molecular graphs (details unveiled in the experiment part). GPS, on the other hand, enjoys a more stable performance and high accuracy at the same time.

\section{Experiment}
The performance of the proposed GPS is evaluated on the real-world molecular datasets for graph classification tasks (e.g., prediction of toxicity of molecular compounds). Also, we present a detailed analysis of the obtained results, aiming to answer the following key questions: \textbf{Q1} (Performance Comparison): Does our proposed method outperform the state-of-the-art baseline methods using the real-world molecular graph dataset? \textbf{Q2} (Sensitivity Analysis): How does the change of sampling parameter $K$ affect the final results? \textbf{Q3} (Ablation Study): To what extent the key component (e.g., correlation coefficient function $\delta$) contribute to the performance of our model?

\subsection{Dataset} 
As the majority of the GNN-based embedding works tend to verify their methods on node classification tasks, where one large graph is given and smaller graphs are then generated from it through mini-batch splitting~\cite{gat,graphsage,graphsaint} (e.g., Cora, Citeseer, Pubmed etc.). We study the graph-level classification task instead, and five different molecular datasets are selected as the benchmark datasets~\cite{wu2018moleculenet}: 1) BBBP records a collections of compounds that whether carry the permeability property to the blood-brain barrier; 2) ClinTox compares FDA approved drugs against terminated drugs due to the toxicity during clinical trials; 3) SIDER involves a list of marketed drugs along with their side effects; 4) BACE collects compounds that acts as the inhibitors of human $\beta$-secretase 1; 5) Tox21 is a widely used public compounds dataset, recording the toxicity of each compound.

As explained in Section 3, we first conduct the pre-processing of the data, which generates the standard input for both our model and other benchmark GNN models (Figure~\ref{fig:demo}(a)), including the initial node feature and adjacency matrix of each graph. Note that GTransformer produces a 8 dimension feature vector for each node in the graph~\cite{rong2020self}. We follow the inductive data setup and randomly split molecular graphs into 80/10/10 partitions for training, validation and test, where the graphs in test partitions remain unobserved throughout the training process. The overview of the characteristics of different datasets\footnote{http://moleculenet.ai/datasets-1} can be seen in Table~\ref{tab:dataset}.
\begin{table}[t!]
\vspace{-1mm}
\centering
\resizebox{0.7\linewidth}{!}{
\begin{tabular}{c||c}
\hline
Name & \ Setup\\
\hline
\hline
Readout func & 2 linear layers, 1 output layer   \\
Loss func & cross entropy loss   \\
Optimizer & Adam~\cite{kingma2014adam}\\
Learning rate & 15e-4 \\
Epochs & 10 \\
Layer depth & 2 \\
Weigh initializer & Xavier~\cite{glorot2010understanding}\\
\hline
\end{tabular}
}
\caption{Shared model setup}
\label{tab:hyperpara}
\vspace{-4mm}
\end{table}
\begin{table}[t!]
\centering
\resizebox{0.9\linewidth}{!}{
\begin{tabular}{c||c|c|c|c}
\hline
Dataset & \# of Graphs & \# of Nodes (Avg.) & \# of Edges (Avg.) & \# of Classes\\
\hline
Tox21 & 7831 & 18.51 & 25.94 & 12   \\
SIDER & 1427 & 33.64 & 35.36 & 27   \\
ClinTox & 1478 & 26.13 & 27.86 & 2  \\
BBBP & 2039 & 24.03 & 25.94 & 1     \\
BACE & 1513 & 34.09 & 36.89 & 1     \\
\hline
\end{tabular}
}
\caption{Summary of molecular datasets}
\label{tab:dataset}
\vspace{-6mm}
\end{table}

\subsection{Experimental Setup} 
\paragraph{Baseline Models}To ensure a fair comparison for predictive tasks on graphs, we re-implement other four baseline models (GCN~\cite{gcn} and GAT~\cite{gat} as conventional GNN models, GraphSAGE~\cite{graphsage} and GraphSAINT~\cite{graphsaint} as sampling-based ones) with the shared model setup using Pytorch\footnote{https://pytorch.org}, including the same layer depth, readout function for graph-level embedding, loss function, Adam optimizer, epochs, and weight initializer. The shared model setup is summarized in Table~\ref{tab:hyperpara}. Note GraphSAINT achieves state-of-the-art results prior our work. In addition, we implement the normalized cosine coefficient (GPS-C), L2 norm distance coefficient (GPS-S) to compare with the default Hadamard product coefficient (GPS) for the ablation study.

\paragraph{Parameter Settings}It is worth mentioning that all baseline models are adapted directly from their officially released code, with the minimum changes possible from their original hyperparameter/architecture setup. As listed in Table~\ref{tab:hyperpara}, we apply the same readout function for all models, with the architecture of shallow MLP as [32, 64, 64]. Note the node embedding size is 32 and final graph embedding size is 64. We select Adam~\cite{kingma2014adam} as the optimizer with the learning rate 15e-4. Also, the layer depth is set to be 2 for all models in the experiment, making it consistent with the best reporting performance for the baseline methods~\cite{graphsage,gcn,gat}. The sample window of GraphSAGE is set to be 3 by default from the released code, and the number of attention heads is 3 in GAT.  While the original GraphSAINT samples sub-GCN graphs from one large graph, we adopt the node sampling from the released source code and choose a smaller node budget and node repeat to meet our experimental scenario (15 and 10, respectively)~\cite{graphsaint}. All trainable parameters of the neural network are initialized through Xavier~\cite{glorot2010understanding}.


\paragraph{Evaluation Metrics}We evaluate the performance of all the models on graph classification task through AUC (i.e. Area Under the ROC Curve~\cite{fawcett2006introduction}), which is widely adopted to multi-label classification scenarios. The multi-label AUC score can be formalized as:
\begin{equation}
    \emph{AUC}_{mul} = \frac{1}{\left\|\emph{y}\right\|}\sum_{\emph{y}_{i}\in\emph{y}}\emph{AUC}\left(\emph{y}_{i}\right)
\end{equation}
where $\emph{AUC}\left(\emph{y}_{i}\right)$ is the area under the curve of ROC for label $\emph{y}_{i}$, and $\left\|\emph{y}\right\|$ represents the total number of label $\emph{y}$. $\emph{AUC}_{mul}$ is the arithmetic average of all labels.

\subsection{Evaluation Results} 
As shown in Table~\ref{tab:hyperpara}, we use a shallow MLP (2 linear layers and 1 output layer) as the readout function to generate final graph-level output, and report the AUC (i.e., Area Under the ROC Curve) scores of all models on graph prediction tasks across five different datasets.
\begin{figure}[t!]
\centering
\includegraphics[width=0.5\textwidth]{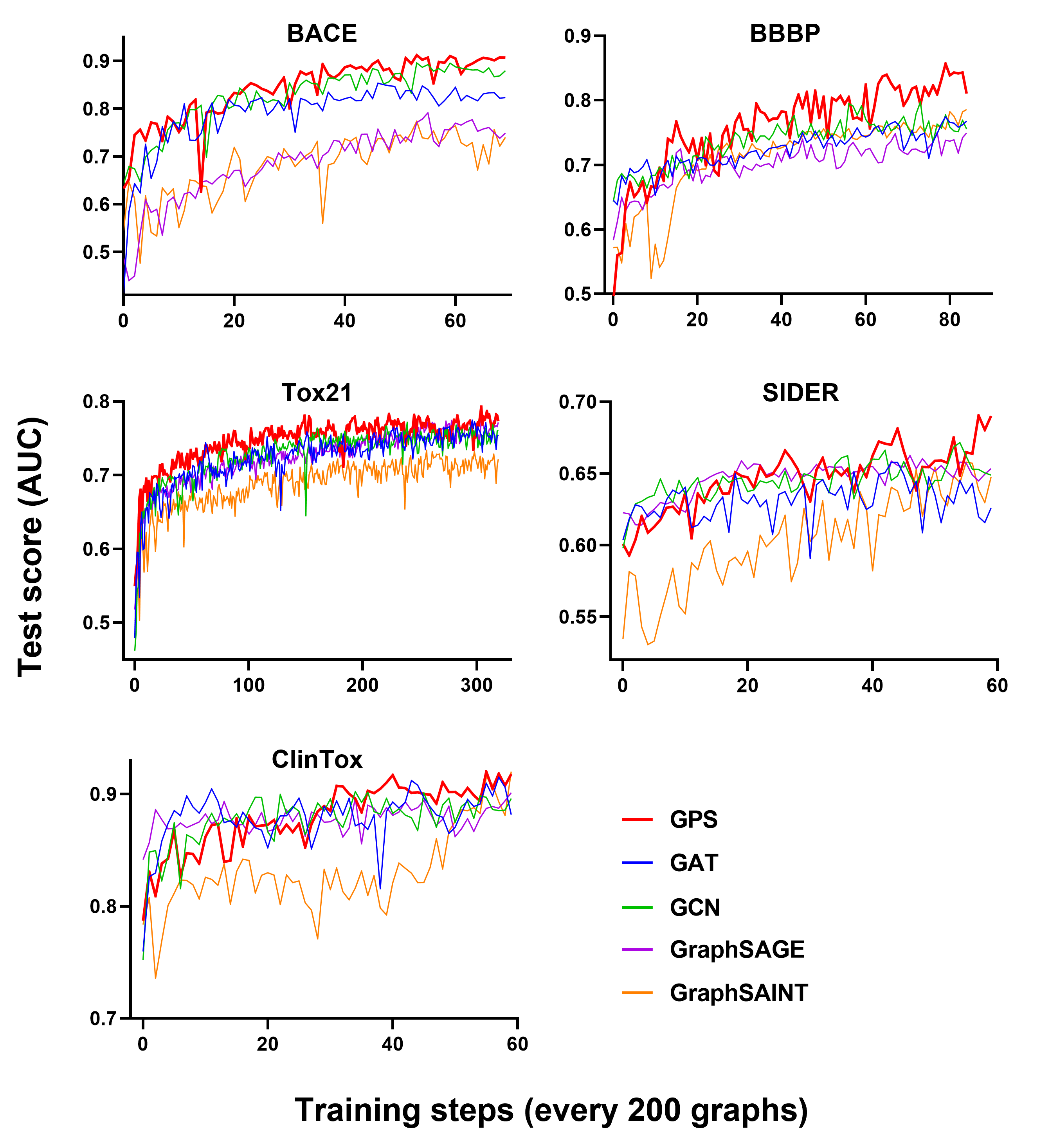}
\vspace{-5mm}
\caption{{Test AUC score at different steps in 10 epochs period. One training epoch is a pass for all graphs in the training set}}
\label{fig:exp_overview}
\vspace{-4mm}
\end{figure}

\paragraph{Performance Comparison (for Q1)} We first look at the performance of our proposed method comparing with other baseline models. Figure~\ref{fig:exp_overview} demonstrates the change of AUC scores from the test set along with the increase of training steps. Note that we record test scores of all unseen graphs for every 200 training graphs as one training step, and the training epoch is set to be 10 for all datasets. We select $K=3$ for our proposed method owing to the reported results, which will be analyzed in detail in the sensitivity analysis part. Compared with four baseline methods, the proposed GPS generally shows higher scores throughout the training.

The best AUC results among all datasets are also recorded and demonstrated in Table~\ref{tbl:bestauc}, in which the highest are marked in bold. Our method outperforms other four baseline models in all datasets in the inductive setting by an average improvement of 4.74\%. In particular, GPS improves the latest reporting sampling model GraphSAINT~\cite{graphsaint} by an average of 6.17\% while demonstrating a much more stable performance throughout the training (as indicated in both Fig.~\ref{fig:exp_overview} and Table~\ref{tbl:bestauc}). Additionally, our method achieves a significant improvement comparing to other conventional (GCN, GAT) and sampling (GraphSAGE) models \textit{w.r.t} the recorded best results, gaining an average increase by 3.7\% on Tox21, 2.8\% on SIDER, 1.07\% on ClinTox, 3.08\% on BBBP, and 8.31\% on BACE respectively. The comparison results with GraphSAGE once again suggest the importance of employing a policy-driven sampling strategy when selecting neighbor nodes rather than using the random sampling seeds.

We also compare our proposed methods with other benchmarks in terms of training speed. As shown in Table~\ref{tbl:time}, GraphSAGE presents the fastest speed owing to less trainable parameters and random sampling strategy. GPS ranks second, showing a small gap with GraphSAGE while significantly improving the AUC score. We also observe splitting a graph into subgraphs in GraphSAINT slows down the training speed, especially for molecular graphs. In general, GPS makes a good balance between training speed and accuracy.

\begin{figure}[t!]
    \centering
    \includegraphics[width=0.85\linewidth]{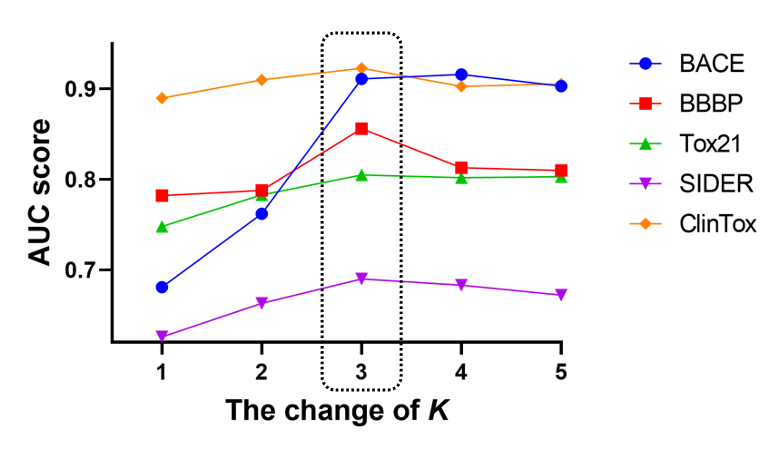}
    \vspace{-5mm}
    \caption{The best AUC score on among different datasets with the change of $K$}
    \label{fig:topk}
    \vspace{-6mm}
\end{figure}

\begin{table}[t!]
\centering
\resizebox{0.5\textwidth}{!}{
\begin{tabular}{|cc|lllll|}
\hline
\multicolumn{2}{|c|}{\multirow{2}{*}{Models}}                                                                & \multicolumn{5}{c|}{Best AUC Scores}                             \\ \cline{3-7} 
\multicolumn{2}{|c|}{}                                                                                        & \multicolumn{1}{l|}{\begin{tabular}[c]{@{}l@{}}Tox21\\ (7831)\end{tabular}} & \multicolumn{1}{l|}{\begin{tabular}[c]{@{}l@{}}SIDER\\ (1427)\end{tabular}} & \multicolumn{1}{l|}{\begin{tabular}[c]{@{}l@{}}ClinTox\\ (1478)\end{tabular}} & \multicolumn{1}{l|}{\begin{tabular}[c]{@{}l@{}}BBBP\\ (2039)\end{tabular}} & \begin{tabular}[c]{@{}l@{}}BACE\\ (1513)\end{tabular} \\ \hline
\multicolumn{1}{|c|}{\multirow{2}{*}{Conventional}} & GCN                                                     & \multicolumn{1}{l|}{0.7714}                                                 & \multicolumn{1}{l|}{0.6715}                                                 & \multicolumn{1}{l|}{0.9021}                                                   & \multicolumn{1}{l|}{0.7974}                                                & 0.8958                                                \\ \cline{2-7} 
\multicolumn{1}{|c|}{}                              & GAT                                                     & \multicolumn{1}{l|}{0.7745}                                                 & \multicolumn{1}{l|}{0.6577}                                                 & \multicolumn{1}{l|}{0.9149}                                                   & \multicolumn{1}{l|}{0.7743}                                                & 0.8534                                                \\ \hline
\multicolumn{1}{|c|}{\multirow{2}{*}{Sampling}}     & GraphSAGE                                               & \multicolumn{1}{l|}{0.7742}                                                 & \multicolumn{1}{l|}{0.6625}                                                 & \multicolumn{1}{l|}{0.9011}                                                   & \multicolumn{1}{l|}{0.7506}                                                & 0.7915                                                \\ \cline{2-7} 
\multicolumn{1}{|c|}{}                              & GraphSAINT                                              & \multicolumn{1}{l|}{0.7337}                                                 & \multicolumn{1}{l|}{0.6579}                                                 & \multicolumn{1}{l|}{0.9198}                                                   & \multicolumn{1}{l|}{0.7856}                                                & 0.7736                                                \\ \hline
\multicolumn{1}{|c|}{\multirow{3}{*}{Our Methods}}         & GPS-D                                                   & \multicolumn{1}{l|}{0.7876}                                                 & \multicolumn{1}{l|}{0.6572}                                                 & \multicolumn{1}{l|}{0.9200}                                                   & \multicolumn{1}{l|}{0.7936}                                                & 0.7840                                                \\ \cline{2-7} 
\multicolumn{1}{|c|}{}                              & GPS-C                                                   & \multicolumn{1}{l|}{0.7927}                                                 & \multicolumn{1}{l|}{0.6869}                                                 & \multicolumn{1}{l|}{0.9130}                                                   & \multicolumn{1}{l|}{0.8285}                                                & 0.8290                                                \\ \cline{2-7} 
\multicolumn{1}{|c|}{}                              & \begin{tabular}[c]{@{}c@{}}\textbf{GPS} (default)\end{tabular} & \multicolumn{1}{l|}{\textbf{0.8005}}                                                 & \multicolumn{1}{l|}{\textbf{0.6905}}                                                 & \multicolumn{1}{l|}{\textbf{0.9202}}                                                   & \multicolumn{1}{l|}{\textbf{0.8566}}                                                & \textbf{0.9117}                                                \\ \hline
\end{tabular}
}
\caption{Best AUC results (higher is better)}
\label{tbl:bestauc}
\vspace{-6mm}
\end{table}
\paragraph{Sensitivity Analysis of Sampling Parameter (for Q2)}We study the impact of sampling parameter $K$ on the overall performance of GPS and explain why we select $K=3$ in the preceding part. As clearly seen in Figure~\ref{fig:topk}, we test different $K$ on all datasets and observe an increasing performance of the model at the beginning. Interestingly, it reaches the best score when $K = 3$ with fluctuation or slight decline afterwards(circled in dashed line). Intuitively, it proves the concept that important neighbor nodes carry most of the topological properties of the graph especially in small molecular graphs, and the traversal to the whole neighborhood may not be needed. This also indicates that the increase of $K$ adds extra noise when updating the node embeddings through message aggregation and thus deteriorate on the unseen graphs in the test set.

\paragraph{Ablation Study (for Q3)}To answer \textbf{Q3}, we implement the variants of GPS, i.e., GPS-D and GPS-C, by replacing the element-wise dot product to L2 norm distance-based and the normalized cosine-based coefficient, all of which are considered as the common strategy when it comes to measuring the correlation scores between vectors~\cite{luo2018cosine}. As indicated in Table~\ref{tbl:bestauc}, the default GPS, which uses the element-wise dot product as the coefficient function in eq.(4), achieves a better performance on all dataset compared with varients GPS-D and GPS-C. Specifically, it improves by an average of 1.03\% on Tox21, 1.84\% on SIDER, 0.37\% on ClinTox, 4.55\% on BBBP, and 10.52\% on BACE. We also observe that the use of simple dot product-based coefficient function requires less computation times when conducting the matrix operations compared to the other two variants.

\begin{table}[t!]
\centering
\resizebox{0.8\linewidth}{!}{
\begin{tabular}{l||lllll}
\hline
Method     & Tox21  & SIDER & ClinTox & BBBP  & BACE  \\ \hline\hline
GCN        & 75.44  & 21.72 & 18.86   & 21.01 & 17.27 \\ \hline
GAT        & 433.03 & 96.72 & 87.82   & 87.12 & 67.76 \\ \hline
GraphSAGE  & 48.12  & 13.41 & 16.13   & 18.55 & 15.09 \\ \hline
GraphSAINT & 337.82 & 75.21 & 65.65   & 68.27 & 54.66 \\ \hline
\textbf{GPS}        & \textbf{69.74}  & \textbf{18.10} & \textbf{16.92}   & \textbf{20.21} & \textbf{17.09} \\ \hline
\end{tabular}
}
\caption{CPU execution time per epoch (second)}
\label{tbl:time}
\vspace{-6mm}
\end{table}
\section{Conclusion}
We present the GPS work in this paper, in which a novel adaptive sampling strategy is designed to generate embeddings effectively for both unseen nodes and graphs. By incorporating the importance-based policy into the sampling process, GPS is able to achieve state-of-the-art performance on graph classification in the inductive setting. The experimental results implicitly prove that our proposed method is able to generalize topological properties of the given graph without knowing the entire graph structure upfront, thus making it easily expandable to larger graphs such as social network or citation network. As future work, an interesting direction is to explore the impact of edge properties in the sampling process, especially for heterogeneous graphs with different edge properties.  
\bibliographystyle{named}
\bibliography{ref}

\begin{thebibliography}{}

\bibitem[\protect\citeauthoryear{Bruna \bgroup \em et al.\egroup
  }{2014}]{bruna2013spectral}
Joan Bruna, Wojciech Zaremba, Arthur Szlam, and Yann LeCun.
\newblock Spectral networks and locally connected networks on graphs.
\newblock In {\em ICLR}, 2014.

\bibitem[\protect\citeauthoryear{Cao \bgroup \em et al.\egroup
  }{2019}]{cao2019multi}
Yixin Cao, Zhiyuan Liu, Chengjiang Li, Juanzi Li, and Tat-Seng Chua.
\newblock Multi-channel graph neural network for entity alignment.
\newblock In {\em ACL}, pages 1452--1461, 2019.

\bibitem[\protect\citeauthoryear{Defferrard \bgroup \em et al.\egroup
  }{2016}]{defferrard2016convolutional}
Micha{\"e}l Defferrard, Xavier Bresson, and Pierre Vandergheynst.
\newblock Convolutional neural networks on graphs with fast localized spectral
  filtering.
\newblock In {\em NeurIPS}, pages 3844--3852, 2016.

\bibitem[\protect\citeauthoryear{Duvenaud \bgroup \em et al.\egroup
  }{2015}]{duvenaud2015convolutional}
David Duvenaud, Dougal Maclaurin, Jorge Aguilera-Iparraguirre, Rafael
  G{\'o}mez-Bombarelli, Timothy Hirzel, Al{\'a}n Aspuru-Guzik, and Ryan~P
  Adams.
\newblock Convolutional networks on graphs for learning molecular fingerprints.
\newblock In {\em NeurIPS}, pages 2224--2232, 2015.

\bibitem[\protect\citeauthoryear{Fawcett}{2006}]{fawcett2006introduction}
Tom Fawcett.
\newblock An introduction to roc analysis.
\newblock {\em Pattern Recognition Letters}, 27(8):861--874, 2006.

\bibitem[\protect\citeauthoryear{Glorot and
  Bengio}{2010}]{glorot2010understanding}
Xavier Glorot and Yoshua Bengio.
\newblock Understanding the difficulty of training deep feedforward neural
  networks.
\newblock In {\em AISTATS}, pages 249--256, 2010.

\bibitem[\protect\citeauthoryear{Grover and
  Leskovec}{2016}]{grover2016node2vec}
Aditya Grover and Jure Leskovec.
\newblock {Node2vec}: Scalable feature learning for networks.
\newblock In {\em SIGKDD}, pages 855--864, 2016.

\bibitem[\protect\citeauthoryear{Hamilton \bgroup \em et al.\egroup
  }{2017}]{graphsage}
William~L Hamilton, Rex Ying, and Jure Leskovec.
\newblock Inductive representation learning on large graphs.
\newblock In {\em NeurIPS}, pages 1025--1035, 2017.

\bibitem[\protect\citeauthoryear{Huang \bgroup \em et al.\egroup
  }{2018}]{huang2018adaptive}
Wenbing Huang, Tong Zhang, Yu~Rong, and Junzhou Huang.
\newblock Adaptive sampling towards fast graph representation learning.
\newblock In {\em ICLR}, page 4563–4572, 2018.

\bibitem[\protect\citeauthoryear{Kingma and Ba}{2015}]{kingma2014adam}
Diederik~P Kingma and Jimmy Ba.
\newblock Adam: A method for stochastic optimization.
\newblock In {\em ICLR}, pages 1269--1272, 2015.

\bibitem[\protect\citeauthoryear{Kipf and Welling}{2017}]{gcn}
Thomas~N Kipf and Max Welling.
\newblock Semi-supervised classification with graph convolutional networks.
\newblock In {\em ICLR}, pages 1--14, 2017.

\bibitem[\protect\citeauthoryear{Liu \bgroup \em et al.\egroup
  }{2019}]{liu2019n}
Shengchao Liu, Mehmet~F Demirel, and Yingyu Liang.
\newblock N-gram graph: Simple unsupervised representation for graphs, with
  applications to molecules.
\newblock {\em NeurIPS}, pages 8466--8478, 2019.

\bibitem[\protect\citeauthoryear{Lou \bgroup \em et al.\egroup
  }{2021}]{lou2021stfl}
Guannan Lou, Yuze Liu, Tiehua Zhang, and Xi~Zheng.
\newblock {STFL}: A temporal-spatial federated learning framework for graph
  neural networks.
\newblock {\em arXiv preprint arXiv:2111.06750}, 2021.

\bibitem[\protect\citeauthoryear{Luo \bgroup \em et al.\egroup
  }{2018}]{luo2018cosine}
Chunjie Luo, Jianfeng Zhan, Xiaohe Xue, Lei Wang, Rui Ren, and Qiang Yang.
\newblock Cosine normalization: Using cosine similarity instead of dot product
  in neural networks.
\newblock In {\em ICANN}, pages 382--391, 2018.

\bibitem[\protect\citeauthoryear{Perozzi \bgroup \em et al.\egroup
  }{2014}]{perozzi2014deepwalk}
Bryan Perozzi, Rami Al-Rfou, and Steven Skiena.
\newblock {Deepwalk}: Online learning of social representations.
\newblock In {\em SIGKDD}, pages 701--710, 2014.

\bibitem[\protect\citeauthoryear{Rong \bgroup \em et al.\egroup
  }{2020a}]{rong2020self}
Yu~Rong, Yatao Bian, Tingyang Xu, Weiyang Xie, Ying Wei, Wenbing Huang, and
  Junzhou Huang.
\newblock Self-supervised graph transformer on large-scale molecular data.
\newblock In {\em NeurIPS}, 2020.

\bibitem[\protect\citeauthoryear{Rong \bgroup \em et al.\egroup
  }{2020b}]{rong2019dropedge}
Yu~Rong, Wenbing Huang, Tingyang Xu, and Junzhou Huang.
\newblock Dropedge: Towards deep graph convolutional networks on node
  classification.
\newblock In {\em ICLR}, 2020.

\bibitem[\protect\citeauthoryear{Veli{\v{c}}kovi{\'c} \bgroup \em et al.\egroup
  }{2018}]{gat}
Petar Veli{\v{c}}kovi{\'c}, Guillem Cucurull, Arantxa Casanova, Adriana Romero,
  Pietro Lio, and Yoshua Bengio.
\newblock Graph attention networks.
\newblock In {\em ICLR}, 2018.

\bibitem[\protect\citeauthoryear{Wang \bgroup \em et al.\egroup
  }{2019}]{wang2019smiles}
Sheng Wang, Yuzhi Guo, Yuhong Wang, Hongmao Sun, and Junzhou Huang.
\newblock {SMILES-BERT}: Large scale unsupervised pre-training for molecular
  property prediction.
\newblock In {\em ACM-BCB}, pages 429--436, 2019.

\bibitem[\protect\citeauthoryear{Weininger \bgroup \em et al.\egroup
  }{1989}]{weininger1989smiles}
David Weininger, Arthur Weininger, and Joseph~L Weininger.
\newblock {SMILES}. 2. algorithm for generation of unique smiles notation.
\newblock {\em Journal of Chemical Information and Computer Sciences},
  29(2):97--101, 1989.

\bibitem[\protect\citeauthoryear{Wu \bgroup \em et al.\egroup
  }{2018}]{wu2018moleculenet}
Zhenqin Wu, Bharath Ramsundar, Evan~N Feinberg, Joseph Gomes, Caleb Geniesse,
  Aneesh~S Pappu, Karl Leswing, and Vijay Pande.
\newblock {MoleculeNet}: A benchmark for molecular machine learning.
\newblock {\em Chemical Science}, 9(2):513--530, 2018.

\bibitem[\protect\citeauthoryear{Wu \bgroup \em et al.\egroup
  }{2020}]{wu2020comprehensive}
Zonghan Wu, Shirui Pan, Fengwen Chen, Guodong Long, Chengqi Zhang, and S~Yu
  Philip.
\newblock A comprehensive survey on graph neural networks.
\newblock {\em IEEE Transactions on Neural Networks and Learning Systems},
  32(1):4--24, 2020.

\bibitem[\protect\citeauthoryear{Ying \bgroup \em et al.\egroup
  }{2018}]{ying2018graph}
Rex Ying, Ruining He, Kaifeng Chen, Pong Eksombatchai, William~L Hamilton, and
  Jure Leskovec.
\newblock Graph convolutional neural networks for web-scale recommender
  systems.
\newblock In {\em SIGKDD}, pages 974--983, 2018.

\bibitem[\protect\citeauthoryear{Yuan \bgroup \em et al.\egroup
  }{2020}]{yuan2020explainability}
Hao Yuan, Haiyang Yu, Shurui Gui, and Shuiwang Ji.
\newblock Explainability in graph neural networks: A taxonomic survey.
\newblock {\em arXiv preprint arXiv:2012.15445}, 2020.

\bibitem[\protect\citeauthoryear{Zeng \bgroup \em et al.\egroup
  }{2020}]{graphsaint}
Hanqing Zeng, Hongkuan Zhou, Ajitesh Srivastava, Rajgopal Kannan, and Viktor
  Prasanna.
\newblock {GraphSAINT}: Graph sampling based inductive learning method.
\newblock In {\em ICLR}, 2020.

\bibitem[\protect\citeauthoryear{Zhang \bgroup \em et al.\egroup
  }{2018}]{zhang2018gaan}
Jiani Zhang, Xingjian Shi, Junyuan Xie, Hao Ma, Irwin King, and Dit-Yan Yeung.
\newblock {GAAN}: Gated attention networks for learning on large and
  spatiotemporal graphs.
\newblock In {\em UAI}, pages 339--349, 2018.

\bibitem[\protect\citeauthoryear{Zhou \bgroup \em et al.\egroup
  }{2020}]{zhou2020graph}
Jie Zhou, Ganqu Cui, Shengding Hu, Zhengyan Zhang, Cheng Yang, Zhiyuan Liu,
  Lifeng Wang, Changcheng Li, and Maosong Sun.
\newblock Graph neural networks: A review of methods and applications.
\newblock {\em AI Open}, 1:57--81, 2020.

\end{thebibliography}

\end{document}